\documentclass[runningheads]{llncs}

\usepackage{eccv}

\usepackage{eccvabbrv}

\usepackage{graphicx}
\usepackage{booktabs}
\usepackage{tabularx}

\usepackage[accsupp]{axessibility}  

\usepackage[pagebackref,breaklinks,colorlinks,citecolor=eccvblue]{hyperref}
\usepackage{hyperref}

\usepackage{orcidlink}

\usepackage{amssymb}
\usepackage{multirow}

\definecolor{codeblue}{rgb}{0.25,0.5,0.5}
\definecolor{nvgreen}{rgb}{0.92, 0.97, 0.85}

\definecolor{navyblue}{HTML}{0071BC}
\definecolor{hotpink}{HTML}{FF0080}

\definecolor{mygreen}{HTML}{FFD500}
\definecolor{myred}{rgb}{1, 0.9, 0.9}
\definecolor{mygray}{gray}{0.95}
\definecolor{mydarkblue}{rgb}{0,0.08,1}
\definecolor{mydarkred}{rgb}{0.8,0.02,0.02}
\definecolor{mydarkorange}{rgb}{0.40,0.2,0.02}
\definecolor{mypurple}{RGB}{111,0,255}
\definecolor{mygold}{rgb}{0.75,0.6,0.12}
\definecolor{mydarkgray}{rgb}{0.66, 0.66, 0.66}
\definecolor{mydarkgreen}{rgb}{0.02,0.6,0.02}
\definecolor{mygray}{gray}{0.9}
\definecolor{keynotegreen}{rgb}{0.04,0.52,0}
\definecolor{keynoteyellow}{rgb}{1,0.68,0}
\definecolor{LightCyan}{rgb}{0.88,1,1}
\definecolor{tabfirst}{rgb}{1, 0.7, 0.7}
\definecolor{tabsecond}{rgb}{1, 0.85, 0.7} 
\definecolor{tabthird}{rgb}{1, 1, 0.7} 
\definecolor{rbtred}{rgb}{255, 0, 0}
\definecolor{mylightgreen}{HTML}{51DA4C}

\usepackage[table]{xcolor}

\newcommand{\methodname}{Cog3DMap\xspace}

\definecolor{firstcolor}{RGB}{245, 190, 50}     
\definecolor{secondcolor}{RGB}{250, 220, 140}   
\definecolor{thirdcolor}{RGB}{254, 242, 200}    

\newcommand{\first}[1]{\cellcolor{firstcolor}\textbf{#1}}
\newcommand{\second}[1]{\cellcolor{secondcolor}{#1}}
\newcommand{\third}[1]{\cellcolor{thirdcolor}{#1}}
\begin{document}

\title{Cog3DMap: Multi-View Vision-Language \\ Reasoning with 3D Cognitive Maps} 

\titlerunning{Cog3DMap}

\author{
Chanyoung Gwak\textsuperscript{*}\inst{1}\orcidlink{0009-0009-6892-4123} \and
Yoonwoo Jeong\textsuperscript{*}\inst{1}\orcidlink{0009-0004-8356-9688} \and
Byungwoo Jeon\inst{2}\orcidlink{0009-0006-9448-2582} \and
Hyunseok Lee\inst{2} \and
Jinwoo Shin\textsuperscript{\textdagger}\inst{2, 3} \and
Minsu Cho\textsuperscript{\textdagger}\inst{1, 3}\orcidlink{0000-0001-7030-1958}
}

\authorrunning{C. Gwak et al.}

\institute{
POSTECH\inst{1}  \,\,\,\,\,\,\, KAIST\inst{2} \,\,\,\,\,\,\, RLWRLD\inst{3}
}

\maketitle
\def\thefootnote{*}\footnotetext{Equal contribution.}\def\thefootnote{\arabic{footnote}}
\def\thefootnote{\textdagger}\footnotetext{Corresponding authors.}\def\thefootnote{\arabic{footnote}}

\begin{abstract}

Precise spatial understanding from multi-view images remains a fundamental challenge for Multimodal Large Language Models (MLLMs), as their visual representations are predominantly semantic and lack explicit geometric grounding.
While existing approaches augment visual tokens with geometric cues from visual geometry models, their MLLM is still required to implicitly infer the underlying 3D structure of the scene from these augmented tokens, limiting its spatial reasoning capability.
To address this issue, we introduce \textbf{\methodname}, a framework that recurrently constructs an explicit 3D memory from multi-view images, where each token is grounded in 3D space and possesses both semantic and geometric information.
By feeding these tokens into the MLLM, our framework enables direct reasoning over a spatially structured 3D map, achieving state-of-the-art performance on various spatial reasoning benchmarks. Code will be made publicly available.
\end{abstract}

\section{Introduction}
\label{sec:main_introduction}

Multimodal Large Language Models (MLLMs)~\cite{li2024llava, chen2024internvl, chen2024internvl25, bai2025qwen25vl, bai2025qwen3} have achieved remarkable success in general visual understanding by leveraging the reasoning capabilities of Large Language Models (LLMs).
Despite their success, precise spatial reasoning from multi-view images remains a fundamental challenge, as their visual representations are predominantly semantic and lack explicit geometric grounding.
To address this limitation, recent studies~\cite{vsibench, fan2025vlm3r} incorporate spatially-annotated training data constructed from ground-truth 3D annotations, while another line of work~\cite{zheng2025vgllm, wu2025spatialmllm, cheng2025sr3d} further augments visual tokens with spatial features extracted from external models such as pointmap estimators~\cite{wang2025vggt}.
Although these approaches enhance the spatial reasoning of MLLMs, they still rely on the MLLM to implicitly infer the underlying 3D structure, without providing an explicit geometric representation of the scene.

A promising direction for overcoming this challenge lies in how humans represent spatial environments.
In cognitive science, this ability is broadly studied under the concept of \emph{spatial memory}, which concerns encoding, storing, and retrieving information about spatial layouts and object locations~\cite{burgess2006spatial, madl2015computational}.
A central framework in this line of research is the \textbf{cognitive map}, an internal representation that encodes spatial relationships among locations in the environment~\cite{tolman1948cognitive, okeefe1978hippocampus}.
Such representations are thought to be acquired incrementally, progressing from recognizing individual landmarks, to learning routes between them, and finally to forming survey-level knowledge that captures metric spatial relationships~\cite{siegel1975development}.

Drawing from this principle, we introduce \textbf{\methodname}, a framework that incrementally constructs a computational counterpart of cognitive maps from multi-view images.
Following the incremental nature of cognitive map formation, \methodname introduces a recurrent framework that progressively integrates multi-view observations into a unified 3D map, where each spatial coordinate is associated with a token carrying both semantic and geometric information.
Unlike prior methods that rely on the MLLM to implicitly infer spatial structure from auxiliary features, \methodname provides the MLLM decoder with an explicit and compact 3D map, enabling more direct and interpretable spatial reasoning.

Experiments on VSTI-Bench~\cite{fan2025vlm3r} and VSI-Bench~\cite{vsibench} demonstrate that our \methodname establishes new state-of-the-art results on spatial reasoning benchmarks.
Moreover, experiments on RoboFAC~\cite{robofac} validate the compactness of \methodname, achieving competitive or superior performance over previous methods while reducing the number of visual tokens by up to 90.2\%.
Furthermore, ablation studies confirm the effectiveness of the proposed explicit 3D map representation in improving spatial understanding. \\

\noindent In summary, our contributions are as follows:
\begin{itemize}
    \item We propose \textbf{\methodname}, a framework that constructs a 3D memory from multi-view images, facilitating spatial understanding of MLLMs through compact and interpretable 3D structure. 

    \item We present an effective strategy to integrate semantic and geometric features for each visual token, thereby enriching it for spatial grounded reasoning. 

    \item We demonstrate state-of-the-art performance across various spatial understanding benchmarks, with ablation studies validating the effectiveness and compactness of using explicit 3D for MLLM spatial reasoning.
\end{itemize}

\section{Related Work}

\subsection{Multimodal Large Language Models (MLLMs)}

Multimodal Large Language Models (MLLMs) have achieved remarkable success across various general visual reasoning tasks by extending the reasoning capabilities of Large Language Models (LLMs) to encompass visual perception. 
For instance, LLaVA~\cite{li2024llava} introduces visual instruction tuning that aligns visual features with the language embedding space via trainable projection layers. 
Building upon this paradigm, subsequent architectures have adopted multi-resolution strategies for high-resolution inputs \cite{liu2024llavanext} or utilized unified multimodal encoders \cite{li2025llavaonevision}.
Among these, the Qwen series \cite{wang2024qwen2, bai2025qwen25vl, bai2025qwen3} has demonstrated exceptional effectiveness across diverse MLLM benchmarks.
Despite these advances, existing MLLMs still exhibit suboptimal performance in spatial understanding tasks requiring geometric interpretation of multi-view images or 3D data, primarily due to the scarcity of spatially-annotated datasets and limited visual grounding capabilities. 
To overcome this limitation, we propose a framework that injects explicit 3D positional information into individual visual tokens, yielding an interpretable and spatially grounded representation that enables effective spatial understanding from multi-view images.

\subsection{Visual Geometry Models (VGMs)}

Predicting 3D structure from multi-view images remains a fundamental challenge in computer vision. 
Traditionally, this problem has been addressed through per-scene optimization via bundle adjustment \cite{schoenberger2016colmap1, schoenberger2016colmap2} or neural field optimization \cite{mildenhall2021nerf, kerbl20233dgs}. 
Recently, VGGT \cite{wang2025vggt} has demonstrated the potential of feed-forward transformers in predicting 3D geometries with high fidelity while eliminating costly per-scene optimizations. 
However, the computational complexity of the transformer restricts scalability to large-scale scenes with numerous viewpoints. 
CUT3R \cite{wang2025cut3r} overcomes this limitation by adopting a recurrent framework that updates the memory state, which is responsible for storing previous observations. 
While this approach improves scalability, it suffers from catastrophic forgetting due to its implicit representation. 
To address this issue, Point3R \cite{wu2025point3r} introduces an explicit token-based memory, where each token stores a 3D position and associated features.
We observe that this recurrent mechanism closely reflects the concept of cognitive maps in cognitive science, progressively constructing a 3D map of the environment as a memory state from sequential observations.

\subsection{Spatial Reasoning of MLLMs}

Early 3D-LLM approaches~\cite{hong20233d, xu2024pointllm, huang2023leo} utilize visual instruction tuning on 3D-text pairs, yet they are constrained by costly 3D data acquisition, resulting in significantly smaller training scales compared to 2D training datasets.
Consequently, the recent research paradigm has shifted toward geometry-augmented multi-view LLMs~\cite{wu2025spatialmllm, zheng2025vgllm, fan2025vlm3r, huang20253drs}, which learn from multi-view images rather than heavily processed 3D data.
In parallel, some works~\cite{fan2025vlm3r} further enhance spatial reasoning by introducing VQA samples automatically generated from metadata such as object dimensions, coordinates, and segmentation masks.
Without requiring explicit 3D inputs at inference time, these models outperform conventional 3D-LLMs on spatial reasoning benchmarks while retaining the broad visual understanding capabilities of general-purpose MLLMs.

A common strategy in this line of work is to leverage VGMs that inject 3D positional information into visual tokens.
For instance, VG-LLM~\cite{zheng2025vgllm}, Spatial-MLLM~\cite{wu2025spatialmllm}, and VLM-3R~\cite{fan2025vlm3r} fuse multi-view visual tokens with geometric features from VGGT~\cite{wang2025vggt}, while 3D-RS~\cite{huang20253drs} distills such cues by aligning final-layer features with VGGT representations.
Concurrent to our work, SR-3D~\cite{cheng2025sr3d} further improves spatial reasoning by enabling flexible region prompting with geometry-augmented features.
Despite their effectiveness, these approaches assign identical 3D coordinates to multiple visual tokens derived from overlapping views, lacking a non-redundant spatial structure that reflects the underlying 3D scene.
This forces the MLLM to disentangle redundant, spatially overlapping information before performing spatial reasoning.
To mitigate this limitation, we introduce \methodname, which constructs a compact 3D token representation where each 3D coordinate is associated with a unique visual token, enabling interpretable and effective spatial reasoning.
\begin{figure}[!t]
    \centering
    \includegraphics[width=1.0\linewidth]{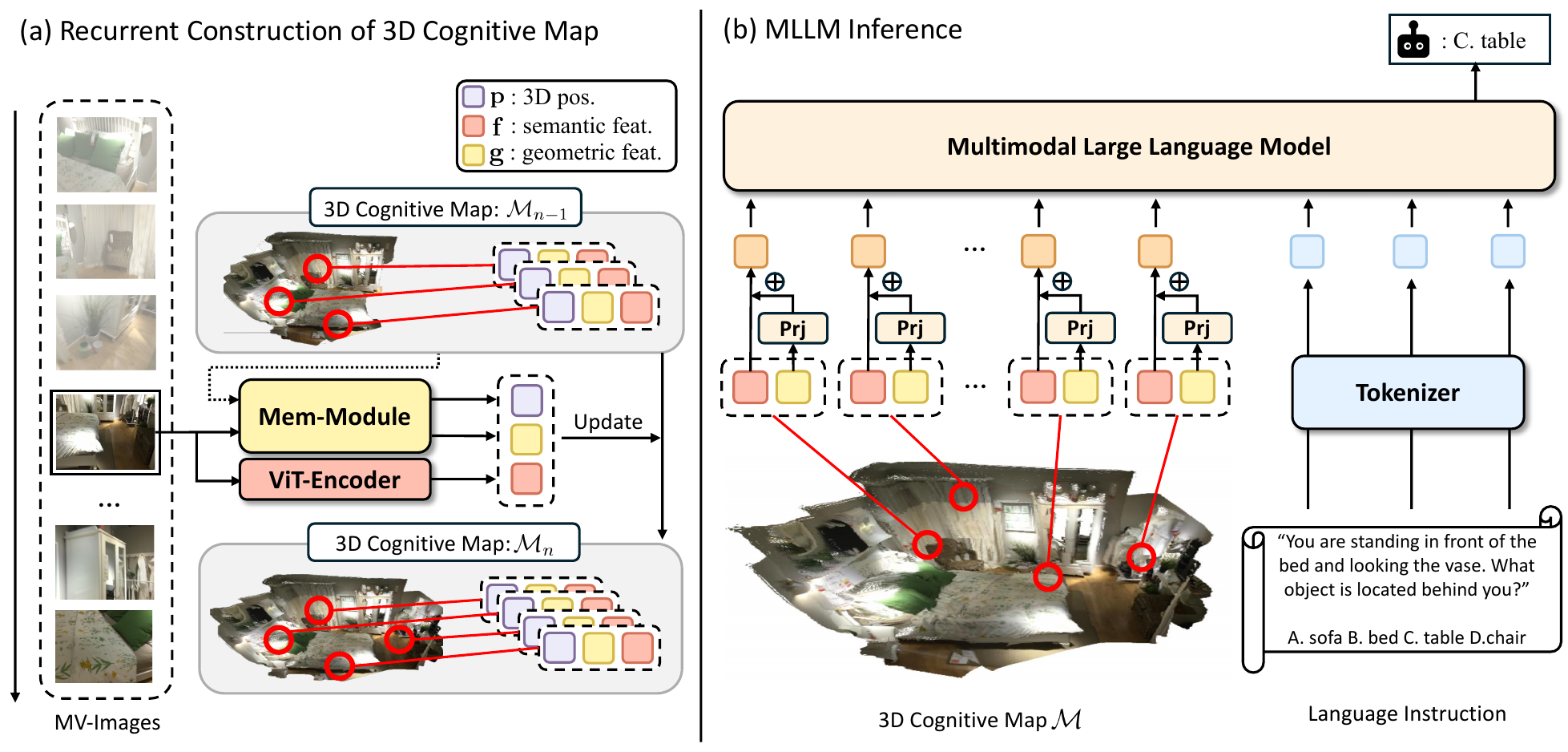}
    \caption{
    Overall pipeline of \methodname. (a) Given a sequence of multi-view images, our recurrent framework, \methodname, progressively integrates visual observations into a unified 3D memory map. Each spatial coordinate in the map is associated with a token carrying both semantic and geometric information. (b) Then, the resulting compact and explicit 3D map is fed into the MLLM decoder for spatial reasoning.
    }
    \label{fig:overall_pipeline}
\end{figure}

\section{Method}

We formulate the task as follows: given a text query $\mathbf{x} = (\mathbf{x}_0, \cdots, \mathbf{x}_{T-1})$ of length $T$ and a set of multi-view images $\mathcal{I} = \{ I_{n} \}_{n=1}^N$, where $N$ is the number of input images, we aim to predict the corresponding natural language response $\mathbf{y}$.
Generating accurate responses in this setting often requires spatial understanding of the underlying 3D scene, \eg, localizing objects or reasoning about their relative arrangements across views.
In this work, we propose \textbf{\methodname}, which leverages an explicit 3D memory $\mathcal{M}$ as an intermediate representation to facilitate spatial understanding of MLLMs.
We first provide an overview in Section~\ref{subsec:overview}, then describe how we construct $\mathcal{M}$ from multi-view images $\mathcal{I}$ in Section~\ref{subsec:spear_description}, and finally elaborate on the implementation details in Section~\ref{subsec:training_details}.

\subsection{Overview: \methodname}
\label{subsec:overview}

We now describe the overall pipeline of \methodname.
As illustrated in Figure~\ref{fig:overall_pipeline}, \methodname (a) recurrently processes multi-view images to construct an explicit 3D memory state $\mathcal{M}$, which we term the \textit{3D Cognitive Map}, and (b) feeds it into an MLLM to generate the final response $\mathbf{y}$.
Since this recurrent framework requires a sequential ordering of the input images $\mathcal{I}$, we follow the dataset-provided order. For tasks that do not require time-dependent reasoning, the image order could also be randomly shuffled.
Initializing from an empty memory state $\mathcal{M}_{0} = \emptyset$, we update the memory state by iterating over all images $\mathcal{I}$.
At each step $n$, upon processing image $I_n$, we update the memory state from $\mathcal{M}_{n-1}$ to $\mathcal{M}_n = \{ (\mathbf{p}_{k}, \mathbf{f}_{k}, \mathbf{g}_{k}) \}_{k=1}^{K_n}$, where each token consists of a 3D position $\mathbf{p}_{k}$, a semantic feature $\mathbf{f}_{k}$, and a geometric feature $\mathbf{g}_{k}$, and $K_n$ denotes the number of memory tokens at step $n$.
The detailed update procedure is described in Section~\ref{subsec:spear_description}.
After processing all images, we obtain the final memory state $\mathcal{M} = \mathcal{M}_N$.
To feed the final memory state $\mathcal{M}$ into the MLLM decoder, we fuse the semantic and geometric features into a single token $\mathbf{v}_{k}$:
\begin{equation}
\mathbf{v}_{k} = \mathbf{f}_{k} + \operatorname{Prj}(\mathbf{g}_{k}),
    \label{eq:feat_update}
\end{equation}
where $\operatorname{Prj}(\cdot)$ denotes learnable projector layers.
Lastly, the MLLM decoder receives the fused visual tokens $\{\mathbf{v}_{k}\}_{k=1}^{K_N}$ and a text query $\mathbf{x}$ to predict the response $\mathbf{y}$.

During training, we optimize the network parameters $\theta$ with a standard cross-entropy loss to maximize the likelihood. Here, $\theta$ comprises the parameters of the projector and the MLLM decoder:
\begin{equation}
    \theta^* = \operatorname{argmax}_{\theta} \; p_{\theta} (\mathbf{y} \mid \{\mathbf{v}_{k}\}_{k=1}^{K}, \mathbf{x}).
\end{equation}

\subsection{Recurrent Construction of 3D Cognitive Map $\mathcal{M}$}
\label{subsec:spear_description}

We now detail how the memory state is updated from $\mathcal{M}_{n-1}$ to $\mathcal{M}_{n}$ upon processing image $I_n$ at step $n$.
For each $I_n$, \methodname proceeds three stages: \textbf{Pointmap Prediction}, \textbf{Semantic Feature Extraction}, and \textbf{Memory Update}, where each stage is detailed below. \\

\noindent{\textbf{Pointmap Prediction.}}
In this stage, a transformer predicts a pointmap $P_n$ and extracts an intermediate geometric feature map $G_n$ from the current image $I_n$, conditioned on the preceding memory state $\mathcal{M}_{n-1}$.
This recurrent design is motivated by recent recurrent pointmap estimators~\cite{wang2025cut3r, wu2025point3r, chen2025ttt3r}.
Specifically, we incorporate the geometric features $\{\mathbf{g}_{k}\}_{k=1}^{K_{n-1}}$ from $\mathcal{M}_{n-1}$ into a pre-trained transformer, where $K_{n-1}$ denotes the number of memory tokens at step $n{-}1$.
Formally, this phase is written as:
\begin{equation}
    (P_n, G_n) =  \operatorname{Mem-Module}(\{\mathbf{g}_{k}\}_{k=1}^{K_{n-1}} , I_n),
    \label{eq:point3r_eq}
\end{equation} 
where $\operatorname{Mem-Module}(\cdot)$ denotes the pre-trained transformer. \\

\noindent{\textbf{Semantic Feature Extraction.}}
In this stage, \methodname extracts a semantic feature map $F_n$ from image $I_n$.
Since the geometric features $G_{n}$ are tailored for 3D reconstruction and lack the semantic information required for language-aligned tasks, we leverage the pre-trained vision encoder of the MLLM to extract complementary semantic features:
\begin{equation}
    F_n = \operatorname{ViT\text{-}Encoder}(I_n),
\end{equation}
where $\operatorname{ViT\text{-}Encoder}(\cdot)$ denotes the pre-trained vision encoder paired with the MLLM decoder.
By reusing the MLLM's own vision encoder, the extracted features are naturally aligned with the language decoder, enabling effective vision-language reasoning without additional alignment training of semantic features.
\\

\noindent{\textbf{Memory Update.}}
After extracting point and feature maps $(P_n, F_n, G_n)$ for image $I_n$, this stage updates the memory state from $\mathcal{M}_{n-1}$ to $\mathcal{M}_n$.
We first compute patch-wise 3D coordinates $\mathbf{p}_{u,v}$ and semantic features $\mathbf{f}_{u,v}$ by averaging over all pixels within each patch $(u, v)$:
\begin{equation}
    \label{eq:patch_point_feat}
    (\mathbf{p}_{u,v}, \mathbf{f}_{u,v}) = \frac{1}{|R_{u,v}|}\sum_{(i, j) \in R_{u,v} } (P_n[i, j],  F_n[i, j])
\end{equation}
where $R_{u,v}$ represents the set of pixel coordinates within patch $(u, v)$ and $[\cdot, \cdot]$ denotes the indexing operation.
For the geometric feature map $G_n$, since it encodes fine-grained structural information that average pooling may discard, we employ an additional encoder to extract patch-wise geometric features:
\begin{equation}
    \mathbf{g}_{u,v} = \operatorname{Encoder}(P_n, G_n) [u,v],
\end{equation}
where $\operatorname{Encoder}(\cdot)$ produces output at the same resolution as the feature maps in Equation~\ref{eq:patch_point_feat}.
We then construct a set of new memory tokens from image $I_n$:
\begin{equation}
\mathcal{M}^{\text{new}}_n= \{ (\mathbf{p}_{u, v}, \mathbf{f}_{u,v}, \mathbf{g}_{u,v}) \mid (u,v) \in \mathcal{P}_n \},
\end{equation}
where $\mathcal{P}_n$ denotes the collection of all patch coordinates in image $I_n$.\\

\noindent We then partition the preceding memory state $\mathcal{M}_{n-1}$ into two disjoint subsets: tokens to be updated $\mathcal{M}_{n-1}^{\text{upd}}$ and tokens to be retained $\mathcal{M}_{n-1}^{\text{ret}}$, such that $\mathcal{M}_{n-1} = \mathcal{M}_{n-1}^{\text{upd}} \cup \mathcal{M}_{n-1}^{\text{ret}}$.
Specifically, for each token $(\mathbf{p}_{k}, \mathbf{f}_{k}, \mathbf{g}_{k}) \in \mathcal{M}_{n-1}$, we compute its minimum 3D distance to the new tokens $\mathcal{M}_n^{\text{new}}$:
\begin{equation}
    d_k = \min_{(\mathbf{p}_{u,v}, \cdot, \cdot) \in \mathcal{M}_n^{\text{new}}} \| \mathbf{p}_{k} - \mathbf{p}_{u,v} \|_2.
\end{equation}
If $d_k < \delta$, where $\delta$ is a pre-defined distance threshold, the token is assigned to $\mathcal{M}_{n-1}^{\text{upd}}$; otherwise, it is assigned to $\mathcal{M}_{n-1}^{\text{ret}}$.
Intuitively, tokens whose 3D positions overlap with the current observation are updated with new information, while tokens far from the current view are retained as-is.
For each token $(\mathbf{p}_{k}, \mathbf{f}_{k}, \mathbf{g}_{k}) \in \mathcal{M}_{n-1}^{\text{upd}}$, we define its neighboring set of new tokens as:
\begin{equation}
    \mathcal{N}_k = \{ (\mathbf{p}_{u,v}, \mathbf{f}_{u,v}, \mathbf{g}_{u,v}) \in \mathcal{M}_n^{\text{new}} \mid \| \mathbf{p}_{k} - \mathbf{p}_{u,v} \|_2 < \delta \}.
    \label{eq:neighbor}
\end{equation}
We then replace each token with the average over $\mathcal{N}_k$, yielding the updated set:
\begin{equation}
    \label{eq:update}
    \hat{\mathcal{M}}_{n-1}^{\text{upd}} = \left\{ \frac{1}{|\mathcal{N}_k|} \sum_{(\mathbf{p}', \mathbf{f}', \mathbf{g}') \in \mathcal{N}_k} (\mathbf{p}', \mathbf{f}', \mathbf{g}') \;\middle|\; (\mathbf{p}_{k}, \mathbf{f}_{k}, \mathbf{g}_{k}) \in \mathcal{M}_{n-1}^{\text{upd}} \right\}.
\end{equation}
We also identify tokens in $\mathcal{M}_n^{\text{new}}$ that do not overlap with any existing memory token:
\begin{equation}
    \mathcal{M}_n^{\text{add}} = \{ (\mathbf{p}_{u,v}, \mathbf{f}_{u,v}, \mathbf{g}_{u,v}) \in \mathcal{M}_n^{\text{new}} \mid \min_{(\mathbf{p}_{k}, \cdot, \cdot) \in \mathcal{M}_{n-1}} \| \mathbf{p}_{u,v} - \mathbf{p}_{k} \|_2 \geq \delta \}.
\end{equation}
These are tokens observing previously unseen regions, which are directly added to the memory.
Finally, we obtain the updated memory state by combining the retained, updated, and newly added tokens:
\begin{equation}
    \mathcal{M}_n = \mathcal{M}_{n-1}^{\text{ret}} \cup \hat{\mathcal{M}}_{n-1}^{\text{upd}} \cup \mathcal{M}_n^{\text{add}}.
\end{equation}
In summary, the memory update mechanism maintains a compact representation of the 3D scene by replacing overlapping tokens with up-to-date observations $\mathcal{M}_{n-1}^{\text{upd}}$, preserving tokens from previously seen regions $\mathcal{M}_{n-1}^{\text{ret}}$, and expanding the memory with tokens from newly observed areas $\mathcal{M}_n^{\text{add}}$.

\subsection{Implementation Details}
\label{subsec:training_details}

We implement our framework based on the official codebase of VG-LLM to ensure a fair comparison with existing models. 
Unless otherwise specified, we adopt Qwen3-VL-8B~\cite{bai2025qwen3} as our MLLM backbone. 
This model strikes an optimal balance between performance across various tasks and computational efficiency. 
For geometric feature extraction and the recurrent pipeline, we employ a pretrained Point3R~\cite{wu2025point3r} and freeze it during training.
Furthermore, we freeze the vision encoder of the MLLM used for semantic feature extraction, while fine-tuning only the decoder.
We also optimize the projector, denoted as $\operatorname{Prj}(\cdot)$ in Equation~\ref{eq:feat_update}, to inject spatial features into the visual tokens.\\

\noindent We observe that certain VQA tasks necessitate temporal reasoning, such as the appearance ordering in VSI-Bench~\cite{vsibench} and tasks in VSTI-Bench~\cite{fan2025vlm3r}.
To resolve this issue, we adopt the video input format of Qwen3-VL, which incorporates temporal separators between timesteps during the processing of multi-view images.
Notably, this design is applicable into other MLLMs by following their format of receiving videos. 
This configuration associates each visual token with specific timestep information, which facilitates temporal understanding.
Furthermore, to ensure training stability, we randomly subsample the memory tokens to a maximum of 8,000 tokens during training, while utilizing the full set of tokens during evaluation. 
Our model is trained on 8 NVIDIA A100 GPUs, requiring approximately 40 GPU hours for the VSI-Bench dataset, while other datasets typically require fewer than 12 hours.
For more implementation details, please refer to Appendix.

\section{Experiments}

We evaluate \methodname on widely used multi-view spatial understanding benchmarks, including VSTI-Bench~\cite{fan2025vlm3r} and VSI-Bench~\cite{vsibench}, in Section~\ref{subsec:vsi_bench}.
In Section~\ref{subsec:robofac}, we additionally evaluate on RoboFAC~\cite{robofac}, a benchmark consisting of question-answer pairs on videos captured in robotic environments, to demonstrate the effectiveness of \methodname on dynamic captures and analyze the compactness of our 3D memory.
Lastly, in Section~\ref{subsec:control_experiments}, we conduct control experiments on Scan2Cap~\cite{chen2021scan2cap}, a widely used 3D-LLM benchmark that requires precise 3D understanding of scenes.
Further details on the evaluation setup and additional results are provided in the Appendix.

\subsection{Evaluation on Spatial Reasoning Benchmarks}
\label{subsec:vsi_bench}

\begin{table}[!t]
    \caption{
        Performance comparison on VSTI-Bench~\cite{fan2025vlm3r}, which evaluates joint spatial and temporal understanding. 
        \textsuperscript{\dag} indicates methods tested on the \texttt{Tiny} subset. 
        \methodname achieves strong performance on spatial reasoning and camera movement prediction tasks, demonstrating its ability to encode both geometric and temporal cues within a unified 3D representation.
        Due to space constraints, we include only the subset of baselines with fewer than 10B parameters. Please refer to Appendix for the full table.
    }
    \label{tab:vstibench}
    \vspace{-2.5mm}
    \centering
    \resizebox{1.00\linewidth}{!}{
    \begin{tabular}{l|c|ccccc}
        & &
        \rotatebox{30}{\textbf{Cam-Obj. Dist.}} &
        \rotatebox{30}{\textbf{Cam. Displce.}} &
        \rotatebox{30}{\textbf{Cam. Mov.}} &
        \rotatebox{30}{\textbf{Obj-Obj. Pose}} &
        \rotatebox{30}{\textbf{Cam-Obj. Dist.}} \\
        \textbf{Model} & \textbf{Avg.} &
        \multicolumn{2}{c}{\cellcolor{orange!20}\textbf{Numerical Answer}} &
        \multicolumn{3}{c}{\cellcolor{yellow!20}\textbf{Multiple-Choice Answer}} \\
        \hline
        \rowcolor{navyblue!5}
        \hline
        \color[HTML]{969696}Random & \color[HTML]{969696}- & \color[HTML]{969696}- & \color[HTML]{969696}- & \color[HTML]{969696}36.1 & \color[HTML]{969696}50.0 & \color[HTML]{969696}36.1 \\
        \color[HTML]{969696}Frequency & \color[HTML]{969696}27.4 & \color[HTML]{969696}5.4 & \color[HTML]{969696}6.2 & \color[HTML]{969696}40.7 & \color[HTML]{969696}52.2 & \color[HTML]{969696}32.4 \\
        \color[HTML]{969696}Human Level\textsuperscript{\dag} & \color[HTML]{969696}77.0 & \color[HTML]{969696}51.4 & \color[HTML]{969696}46.8 & \color[HTML]{969696}95.1 & \color[HTML]{969696}97.5 & \color[HTML]{969696}94.3 \\
        \hline
        \rowcolor{navyblue!5}
        \multicolumn{1}{l|}{\textcolor{black}{\textit{Proprietary Models (API)}}} & & & & & & \\
        Gemini-1.5-Flash~\cite{team2024gemini} & 32.1 & 28.5 & 20.9 & 24.4 & 52.6 & 33.9 \\
        GPT-4o~\cite{openai2024gpt4o} & 38.2 & 29.5 & 23.4 & 37.3 & 58.1 & 42.5 \\
        \hline
        \rowcolor{navyblue!5}
        \multicolumn{1}{l|}{\textcolor{black}{\textit{Open-source Models}}} & & & & & & \\
        LongVILA-8B~\cite{xue2024longvila} & 30.5 & 20.0 & 11.6 & 35.4 & 52.3 & 33.4 \\
        LongVA-7B & 32.3 & 13.5 & 5.1 & 43.7 & 57.9 & 41.2 \\
        VILA-1.5-8B~\cite{lin2024vila} & 37.3 & 30.1 & \third{27.3} & 42.2 & 50.4 & 36.7 \\
        LLaVA-NeXT-Video-7B~\cite{liu2024llavanext} & 40.0 & 28.2 & 1.8 & \third{49.8} & 64.7 & \third{55.6} \\
        LLaVA-OneVision-7B~\cite{li2025llavaonevision} & 41.7 & 29.9 & 19.3 & 47.5 & 62.1 & 49.8 \\
        InternVL2-8B~\cite{chen2024internvl} & \third{43.5} & \third{32.9} & 13.5 & 48.0 & \third{68.0} & 55.0 \\
        \hline
        \rowcolor{navyblue!5}
        \multicolumn{1}{l|}{\textcolor{black}{\textit{Spatial-Enhanced Models}}} & & & & & & \\
        VLM-3R-7B~\cite{fan2025vlm3r} & \second{58.8} & \second{39.4} & \second{39.6} & \second{60.6} & \second{86.5} & \second{68.6} \\
        \methodname-8B (ours) & \first{67.5} & \first{40.9} & \first{47.1} & \first{88.1} & \first{90.9} & \first{70.6} \\
        \bottomrule
    \end{tabular}
    }
\vspace{-3.5mm}
\end{table}

We evaluate our \methodname on VSTI-Bench~\cite{fan2025vlm3r}, a benchmark that primarily focuses on spatial questions requiring complex understanding of camera and object locations.
In particular, VSTI-Bench incorporates temporal context through tasks such as action-spatial grounding, trajectory prediction, and temporal-spatial relationship reasoning.
For training, we use the official training split of VSTI-Bench and follow the official evaluation protocol.
Notably, we preserve the sample ordering provided by the official dataset to ensure correct evaluation.

As reported in Table~\ref{tab:vstibench}, our \methodname shows consistent gains across all tasks over the previous state-of-the-art model, VLM-3R-7B~\cite{fan2025vlm3r}, achieving an 8.7$\%p$ improvement in the average score.
In particular, camera movement prediction shows the largest improvement, a $27.5\%p$ improvement, among all tasks.
Note that VLM-3R~\cite{fan2025vlm3r} uses tokens obtained by CUT3R~\cite{wang2025cut3r}, which are fixed-length tokens that store scene information implicitly.
In contrast, our \methodname preserves the explicit 3D spatial structure of multi-view images through dedicated 3D visual tokens.
This design allows our 3D tokens to encode both geometric and semantic cues alongside temporal information within a unified representation, facilitating joint spatial and temporal understanding.

\begin{table}[t]
    \caption{Results on multi-view global spatial scene understanding on VSI-Bench~\cite{yang2024thinking}. \textit{Spatial-Enhanced Models} denote methods specialized for spatial reasoning. \methodname achieves state-of-the-art overall performance. It performs particularly well on absolute distance, relative direction, and appearance order, demonstrating the benefits of explicit 3D representations for spatial reasoning. 
    }
    \label{tab:3d_vsibench}
    \vspace{-2.5mm}
    \centering
    \resizebox{1.00\linewidth}{!}{
    \begin{tabular}{l|c|cccccccc}
        & & 
        \rotatebox{30}{\textbf{Obj. Count}} &
        \rotatebox{30}{\textbf{Abs. Dist.}} &
        \rotatebox{30}{\textbf{Obj. Size}} & 
        \rotatebox{30}{\textbf{Room Size}} &
        \rotatebox{30}{\textbf{Rel. Dist.}} &
        \rotatebox{30}{\textbf{Rel. Dir.}} &
        \rotatebox{30}{\textbf{Route Plan}} &
        \rotatebox{30}{\textbf{Appr. Order}} \\
        \textbf{Model} & \textbf{Avg.} & \multicolumn{4}{c}{\cellcolor{orange!20}\textbf{Numerical Answer}} & \multicolumn{4}{c}{\cellcolor{yellow!20}\textbf{Multiple-Choice Answer}} \\
        \hline
        \rowcolor{navyblue!5}
        \hline
        \color[HTML]{969696}Random & \color[HTML]{969696}  & \color[HTML]{969696} & \color[HTML]{969696}- & \color[HTML]{969696}- & \color[HTML]{969696}- & \color[HTML]{969696}25.0 & \color[HTML]{969696}36.1 & \color[HTML]{969696}28.3 & \color[HTML]{969696}25.0 \\ 
        \color[HTML]{969696}Human Level\textsuperscript{\dag} & \color[HTML]{969696}79.2 & \color[HTML]{969696}94.3 & \color[HTML]{969696}47.0 & \color[HTML]{969696}60.4 & \color[HTML]{969696}45.9 & \color[HTML]{969696}94.7 & \color[HTML]{969696}95.8 & \color[HTML]{969696}95.8 & \color[HTML]{969696}100.0 \\
        \hline
        \rowcolor{navyblue!5}
    
        \multicolumn{1}{l|}{\textcolor{black}{\textit{Proprietary Models (API)}}} & & & & & & & & & \\
        GPT-4o~\cite{openai2024gpt4o} & 34.0 & 46.2 & 5.3 & 43.8 & 38.2 & 37.0 & 41.3 & 31.5 & 28.5 \\
        Gemini-1.5-Flash~\cite{team2024gemini} & 42.1 & 49.8 & 30.8 & 53.5 & {54.4} & 37.7 & 41.0 & 31.5 & 37.8 \\
        Gemini-1.5-Pro~\cite{team2024gemini} & 45.4 & {56.2} & {30.9} & 64.1 & 43.6 & {51.3} & 46.3 & {36.0} & 34.6 \\
        \hline
        \rowcolor{navyblue!5}
        \multicolumn{1}{l|}{\textcolor{black}{\textit{Open-source Models}}} & & & & & & & & & \\
        VILA-1.5-8B~\cite{lin2024vila} & 28.9 & 17.4 & 21.8 & 50.3 & 18.8 & 32.1 & 34.8 & 31.0 & 24.8 \\
        LLaVA-OneVision-7B~\cite{li2025llavaonevision}  & 32.4 & 47.7 & 20.2 & 47.4 & 12.3 & 42.5 & 35.2 & 29.4 & 24.4 \\
        InternVL2-8B~\cite{chen2024internvl} & 34.6 & 23.1 & {28.7} & 48.2 & {39.8} & 36.7 & 30.7 & 29.9 & 39.6 \\
        LLaVA-NeXT-Video-7B~\cite{zhang2024llavanextvideo} & 35.6 & 48.5 & 14.0 & 47.8 & 24.2 & {43.5} & 42.4 & 34.0 & 30.6 \\
        \hline
        \rowcolor{navyblue!5}
        \multicolumn{1}{l|}{\textcolor{black}{\textit{Spatial-Enhanced Models}}} & & & & & & & & & \\

        VG-LLM-4B~\cite{zheng2025vgllm} & 47.3 & 66.0 & 37.8 & 55.2 & 59.2 & 44.6 & 45.6 & 33.5 & 36.4 \\
        Spatial-MLLM-4B~\cite{wu2025spatialmllm} & 48.4 & 65.3 & 34.8 & 63.1 & 45.1 & 41.3 & 46.2 & 33.5 & 46.3 \\
        VG-LLM-8B~\cite{zheng2025vgllm} & 50.7 & 67.9 & 37.7 & 58.6 & 62.0 & 46.6 & 40.7 & 32.4 & 59.2\\
        3DRS-7B~\cite{huang20253drs} & 45.9 & 68.7 & 34.8 & 53.6 & 56.6 & 40.9 & 43.2 & 30.4 & 39.2 \\
        VLM-3R-7B~\cite{fan2025vlm3r} & \third{60.9} & \second{70.2} & \second{49.4} & \second{69.2} & \second{67.1} & \first{65.4} & \second{80.5} & \first{45.4} & \third{40.1} \\
        VST-7B~\cite{yang2025vst} & \second{61.2} & \first{71.6} & \third{43.8} & \first{75.5} & \first{69.2} & \third{60.0} & \third{55.6} & \second{44.3} & \first{69.2} \\
        \methodname-8B (ours) & \first{65.1} & \third{69.6} & \first{54.8} & \third{67.8} & \second{67.1} & \second{64.8} & \first{85.6} & \third{43.0} & \second{67.9} \\
        \bottomrule
    \end{tabular}
    }
    \vspace{-3.5mm}
\label{tab:vsibench}
\end{table}

We also compare \methodname with previous approaches on VSI-Bench~\cite{vsibench}, a widely used benchmark that evaluates the spatial awareness of MLLMs through questions focused on spatial grounding, spatial relationship reasoning, and size estimation.
As shown in Table~\ref{tab:3d_vsibench}, our model establishes a new state-of-the-art by outperforming the previous leading model, VST-7B~\cite{yang2025vst}, with an average improvement of $3.9\%p$.
We attribute this improvement to key differences in design.
Specifically, VST-7B introduces additional training samples coupled with reinforcement learning, requiring the MLLM to implicitly learn spatial understanding capabilities.

Furthermore, \methodname achieves superior performance relative to models leveraging visual geometry foundation models~\cite{zheng2025vgllm, wu2025spatialmllm, fan2025vlm3r}.
Previous approaches frequently struggle with absolute distance estimation, particularly when multiple redundant tokens represent the same spatial area.
Under these conditions, the MLLM struggles to identify the specific token corresponding to the target object, complicating the reasoning process.
In contrast, our framework constructs a compact, non-redundant 3D map where each spatial location is represented by a unique token, facilitating more accurate and efficient spatial reasoning.

\subsection{Evaluation on the RoboFAC Benchmark}
\label{subsec:robofac}

\begin{figure}[t]
  \centering
    \includegraphics[width=\linewidth]{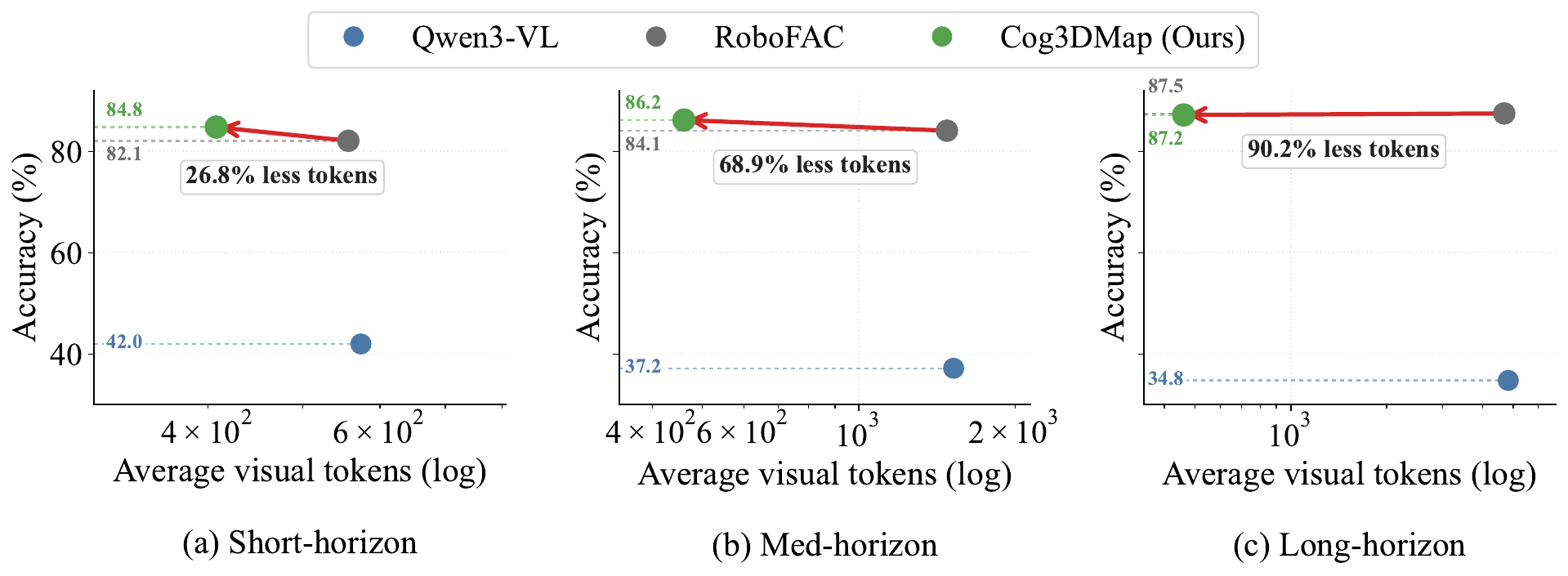}
    \vspace{-0.25in}
  \caption{We compare accuracy (\%) against the average number of visual tokens (log scale) for short, medium, and long horizon tasks. \methodname(Ours) consistently dominates the trade-off, achieving comparable or better accuracy while requiring substantially fewer visual tokens (up to 90.2\% reduction), demonstrating improved token efficiency across horizons.
  }
  \label{fig:robofac}
\end{figure}

We evaluate \methodname on RoboFAC~\cite{robofac}, a dataset consisting of videos that capture dynamic interactions between robot agents and objects.
Specifically, we evaluate on `short-horizon', `medium-horizon', and `long-horizon' tasks, each consisting of short, medium, and long video sequences with corresponding question-answer samples.
In this experiment, we assess the efficiency of our framework by comparing \methodname with Qwen3-VL trained on RoboFAC with respect to the number of tokens and performance.
By evaluating on videos with dynamic actions, we also showcase the applicability of our model to dynamic video sequences.
We use Qwen3-VL-4B as the backbone network and train both our model and Qwen3-VL-4B with the same hyperparameters on the official training set of RoboFAC.
Following the naming convention in the original RoboFAC paper, we denote Qwen3-VL-4B trained on RoboFAC as RoboFAC-4B.
We additionally report the performance of Qwen3-VL-4B without additional training, denoted as Qwen3-VL-4B in Figure~\ref{fig:robofac}.
We report the average success rate evaluated by the publicly available Qwen3 model, along with the average number of tokens across all samples in the evaluation set.
For further details on training and evaluation, we refer the reader to the Appendix.

As shown in Figure~\ref{fig:robofac}, \methodname achieves substantial reductions in the number of visual tokens across all three task horizons while maintaining competitive or superior performance.
In the short-horizon setting, \methodname outperforms RoboFAC-4B by 2.7\%p with 26.8\% fewer tokens.
The efficiency gain becomes more pronounced in the medium-horizon setting, where \methodname surpasses RoboFAC-4B by 2.1\%p while requiring 68.9\% fewer tokens.
In the long-horizon setting, \methodname nearly matches RoboFAC-4B with a 90.2\% reduction in the number of tokens.
These results demonstrate that \methodname effectively compresses visual information without sacrificing task performance, with the token efficiency gain being particularly pronounced for longer video sequences.

Furthermore, the competitive performance on RoboFAC demonstrates the applicability of \methodname to dynamic scenes that require spatio-temporal understanding.
The spatial neighborhood-based feature update in Equation~\ref{eq:update}, however, is a suboptimal workaround for highly dynamic scenes.
Since features at each coordinate are aggregated from nearby tokens, the trajectory of an object is overwritten when another object subsequently passes through the same spatial region, making it challenging to maintain distinct motion histories across multiple objects.
Developing a more effective update strategy for such scenarios remains an open direction for future work.

\subsection{Control Experiments}
\label{subsec:control_experiments}

We conduct several control experiments to validate the effectiveness of each component of our proposed model. 
All training and evaluation are performed on the Scan2Cap~\cite{chen2021scan2cap} dataset, which requires precise spatial understanding, as each sample consists of a description of an object at a specific coordinate. 
To ensure a rigorous comparison with previous multi-view LLMs, we follow the evaluation protocol provided by VG-LLM~\cite{zheng2025vgllm}, aligning the predicted 3D map with the GT point cloud, since the evaluation samples assume the GT point cloud as given. 
For additional control experiments, please refer to the Appendix.\\

\begin{table}[!t]
    \caption{Ablation study on positional embedding strategies evaluated on Scan2Cap~\cite{chen2021scan2cap}. Learnable-PE adopts learnable Fourier bases, 4D-RoPE replaces the native 3D-RoPE in Qwen3, and HRoPE employs hierarchical positional embeddings. For each memory token: $\mathbf{f}_k$ denotes the semantic features, $\mathbf{g}_k$ denotes the geometric features, and $\mathbf{p}_k$ denotes the 3D coordinate. \textbf{C}, \textbf{B-4}, \textbf{M}, and \textbf{R} denote CIDEr, BLEU-4, METEOR, and ROUGE-L at an IoU threshold of 0.5.}
    \label{tab:pos_emb}
    \small
    \centering
    \vspace{-2mm}
    \setlength{\tabcolsep}{10pt}
    \resizebox{0.90\linewidth}{!}{
    \begin{tabular}{l|cccc}
    \toprule
    \textbf{Model} & \textbf{C$\uparrow$} & \textbf{B-4$\uparrow$} & \textbf{M$\uparrow$} & \textbf{R$\uparrow$} \\
    \midrule
    Qwen3-VL-8B~\cite{bai2025qwen3} & 65.2 & 37.8 & 27.2 & 61.1 \\
    \midrule
    $\mathbf{f}_k + \operatorname{Learnable-PE}(\mathbf{p}_k)$  & 66.7 & 37.9 & 27.2 & 61.1 \\
    $\mathbf{f}_k + \operatorname{4D-RoPE}(\mathbf{p}_k)$ & 64.0 & 37.4 & 27.1 & 61.0 \\
    $\mathbf{f}_k + \operatorname{HRoPE}(\mathbf{p}_k)$ & 71.7 & 39.0 & 27.8 & 61.6 \\
    $\mathbf{f}_k + \operatorname{Prj}(\mathbf{g}_k)$ & \textbf{79.4} & \textbf{40.8} & \textbf{28.4} & \textbf{62.3} \\
    $\mathbf{f}_k + \operatorname{Prj}(\mathbf{g}_k) + \operatorname{Learnable-PE}(\mathbf{p}_k)$ & 78.0 & 40.2 & 28.3 & 62.2 \\
    $\mathbf{f}_k + \operatorname{Prj}(\mathbf{g}_k) + \operatorname{4D-RoPE}(\mathbf{p}_k)$  & 77.9 & 40.5 & 28.4 & 62.4 \\
    $\mathbf{f}_k + \operatorname{Prj}(\mathbf{g}_k) + \operatorname{HRoPE}(\mathbf{p}_k)$ & 76.6 & 39.9 & 28.2 & 61.8 \\
    \bottomrule
    \end{tabular}
    }
\end{table}
\noindent{\textbf{Positional Embedding.}} 
Each token in the memory state $\mathcal{M} = \{ (\mathbf{p}_k, \mathbf{f}_k, \mathbf{g}_k) \}$ comprises a 3D position $\mathbf{p}_k$, semantic features $\mathbf{f}_k$, and geometric features $\mathbf{g}_k$.
We explore several fusion designs to combine these features into a single visual token.
For fusing $\mathbf{p}_k$ with $\mathbf{f}_k$, we evaluate two positional embedding strategies: Learnable PE and 4D-RoPE.
4D-RoPE extends the native 3D-RoPE$(t, h, w)$ of Qwen3-VL to a 4D format $(t, x, y, z)$.
Learnable PE composes input positions $\mathbf{p}_k$ with learnable basis frequencies.
We additionally evaluate the hierarchical positional embedding ($\operatorname{HRoPE}$) from Point3R~\cite{wu2025point3r}, which encodes positions through multiple frequency bands.
We further investigate incorporating geometric features $\mathbf{g}_k$ to enrich the representation for scene reconstruction.
A learnable projector maps geometric features $\mathbf{g}_k$ into the semantic feature space of $\mathbf{f}_k$.

As illustrated in Table~\ref{tab:pos_emb}, incorporating geometric features $\mathbf{g}_k$ through a projector yields the most significant improvement, boosting C@0.5 from 65.2 to 79.4 (+14.2) over the baseline that relies solely on semantic features $\mathbf{f}_k$.
Among positional embedding strategies applied without geometric features, HRoPE achieves the highest gain (71.7 C@0.5), followed by Learnable-PE (66.7), while 4D-RoPE degrades performance below the baseline (64.0).
We attribute this degradation to interference between 4D-RoPE and the native 3D-RoPE of Qwen3-VL, which already encodes spatial patterns through $(h, w)$ image coordinates.
Appending any explicit 3D positional embedding on top of geometric features consistently lowers performance relative to using geometric features alone (79.4 $\rightarrow$ 78.0, 77.9, 76.6 in C@0.5).
This indicates that geometric features from Point3R already encode sufficient spatial information, and additional 3D coordinate embeddings become redundant.
Therefore, \methodname adopts the fusion of $\mathbf{f}_k$ and $\operatorname{Projector}(\mathbf{g}_k)$ as the default configuration.
\\

\begin{table}[!t]
    \caption{Comparison of 3D information injection strategies on dense video captioning. All variants adopt the same backbone architecture. \textbf{C}, \textbf{B-4}, \textbf{M}, and \textbf{R} denote CIDEr, BLEU-4, METEOR, and ROUGE-L, respectively, evaluated at an IoU threshold of 0.5.}
    \label{tab:model_design}
    \small
    \centering
    \setlength{\tabcolsep}{8pt}
    \resizebox{1.0\linewidth}{!}{
    \begin{tabular}{l|cccc|cccc}
    \toprule
    \multirow{2}{*}{\textbf{Model}} & \multicolumn{4}{c|}{Qwen3-VL-4B} & \multicolumn{4}{c}{Qwen3-VL-8B} \\
    \cmidrule(lr){2-5} \cmidrule(lr){6-9}
    & \textbf{C$\uparrow$} & \textbf{B-4$\uparrow$} & \textbf{M$\uparrow$} & \textbf{R$\uparrow$}  & \textbf{C$\uparrow$} & \textbf{B-4$\uparrow$} & \textbf{M$\uparrow$} & \textbf{R$\uparrow$} \\
    \midrule
    3D-REPA~\cite{yu2024representation, huang20253drs} & 31.2 & 10.5 & 18.3 & 45.7 & 32.2 & 10.9 & 18.3 & 46.0 \\
    VGM-Aug~\cite{zheng2025vgllm, wu2025spatialmllm, fan2025vlm3r} & 58.3 & 32.6 & 26.2 & 60.3 & 59.5 & 32.8 & 26.2 & 60.3 \\
    \methodname (Ours) & \textbf{75.9} & \textbf{39.7} & \textbf{28.0} & \textbf{61.8} & \textbf{79.4} & \textbf{40.8} & \textbf{28.4} & \textbf{62.3} \\
    \bottomrule
    \end{tabular}
    }
\end{table}

\noindent{\textbf{3D Feature Integration Strategy.}}
In this experiment, we compare various strategies for injecting spatial understanding capabilities from multi-view images into MLLMs.
The first variant, denoted as 3D-REPA, applies representation alignment with multi-view images as introduced in 3DRS~\cite{huang20253drs}.
This approach aligns the internal representations of the MLLM with geometric features extracted by visual geometry models, allowing the model to implicitly acquire spatial understanding during training.
At inference, 3D-REPA does not require geometric features from the foundation models.
The second variant, denoted as VGM-Aug, directly augments visual tokens by injecting geometric features from visual geometry models.
Unlike 3D-REPA, VGM-Aug relies on the visual geometry models at inference as well.
We compare both variants with \methodname in Table~\ref{tab:model_design}.
For fair comparison, all variants adopt the same backbone architecture, Qwen3-VL-4B and Qwen3-VL-8B.

As shown in Table~\ref{tab:model_design}, \methodname consistently outperforms both variants across all metrics on both backbone architectures.
3D-REPA~\cite{yu2024representation, huang20253drs} yields the lowest performance among all variants, obtaining a CIDEr score of 31.2 on Qwen3-VL-4B.
Without access to explicit geometric information at inference, the model relies entirely on the implicit alignment signal, which alone does not provide sufficient guidance for fine-grained spatial understanding.
VGM-Aug~\cite{zheng2025vgllm, wu2025spatialmllm, fan2025vlm3r} significantly improves over 3D-REPA by incorporating explicit geometric features from visual geometry models, achieving a CIDEr score of 58.3 on Qwen3-VL-4B.
However, such geometric features are directly appended to the visual token sequence without compression.
Since all visual tokens participate in the attention mechanism of the MLLM, multiple tokens originating from overlapping viewpoints occupy the same spatial location, each carrying redundant information.
Even with 3D positional information, this redundancy introduces an additional challenge for the MLLM to disentangle spatially overlapping tokens and extract meaningful representations from the duplicated observations.

In contrast, \methodname constructs a compact 3D memory map that eliminates redundant information through spatial aggregation.
The resulting memory tokens provide an interpretable and compact representation of the 3D scene, allowing the MLLM to perform spatial reasoning without the burden of processing redundant observations.
This reduced complexity in the input representation lowers the difficulty of spatial understanding for the MLLM, leading to consistent performance gains across both model scales, with \methodname achieving CIDEr scores of 75.9 and 79.4 on Qwen3-VL-4B and Qwen3-VL-8B, respectively. \\

\begin{figure}[!t]
    \centering
    \includegraphics[width=0.98\linewidth]{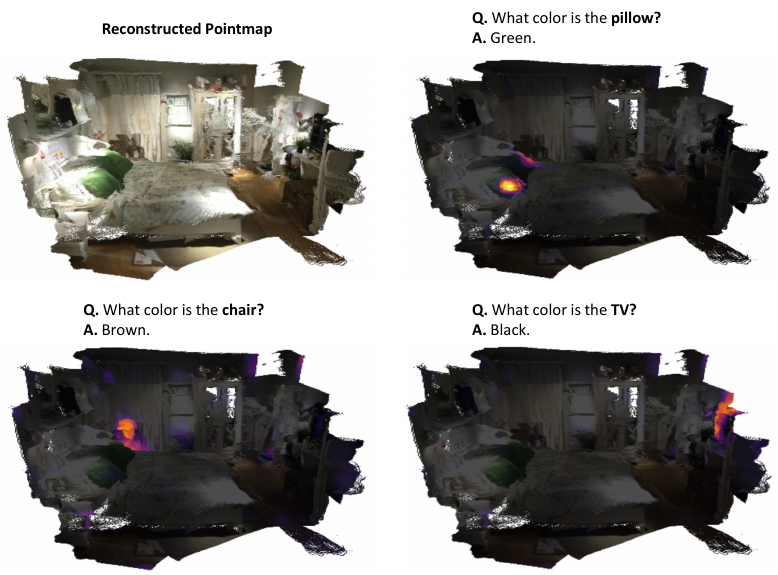}
    \caption{Visualization of attention scores over visual tokens across varying text queries. To analyze the model behavior, we fix the scene and vary the target object in the text query. Our \methodname assigns high attention scores to visual tokens relevant to the query object, without explicit supervision for such attention. }
    \label{fig:attn_viz_main}
\end{figure}

\noindent{\textbf{Qualitative Analysis.}}
To analyze the behavior of \methodname, we visualize the attention maps of visual tokens with respect to text queries.
Following standard attention visualization practice, we zero out the attention weights of sink tokens to prevent them from dominating the visualization, and highlight regions with high attention scores.
As shown in Figure~\ref{fig:attn_viz_main}, we present attention maps from the same scene \texttt{scene0706\_00} with varying text queries.
For each token, the 3D position is plotted based on coordinates predicted by Point3R~\cite{wu2025point3r} for visualization purposes.
\methodname assigns high attention weights to tokens situated near the query location, without explicit supervision for spatial alignment.
Notably, the high-attention tokens form spatially coherent clusters in 3D space rather than appearing as scattered points across the scene.
This coherence demonstrates that \methodname acquires robust 3D spatial understanding, driven by the spatial grounding of the memory map and the rich semantic features of Qwen's vision encoder.

\section{Conclusion}

In this work, we presented \methodname, an external memory-based framework that enhances the spatial reasoning capabilities of MLLMs through explicit 3D representations.
\methodname transforms multi-view images into compact 3D tokens via precise coordinate alignment, reducing redundancy from overlapping visual signals while preserving essential scene context.
Extensive evaluations on challenging benchmarks demonstrated that \methodname achieves state-of-the-art performance in spatial reasoning tasks, validating the effectiveness of integrating explicit 3D grounding within large-scale vision-language models.
Moreover, our control experiments validated the compactness of the 3D memory state over long-horizon sequences.
These contributions open a new direction for scene-level spatial understanding without relying on dense pixel-level computations. \\

\noindent{\textbf{Limitations.}}
Although \methodname achieves strong performance across various benchmarks, we have not yet validated the framework in highly dynamic scenes.
We observe that the recurrent framework encounters difficulty when multiple objects pass through the same spatial region, where subsequent aggregation overwrites features at shared coordinates.
We leave the extension of our pipeline to extremely dynamic scenes and the development of corresponding training and evaluation datasets for future work.
Furthermore, our proposed pipeline requires two-stage training, as the geometric features need to be optimized prior to training the LLM.
Developing a fully end-to-end training pipeline remains a promising direction for more robust spatial understanding from multi-view images.
We provide further discussion on limitations and future work in the Appendix.

\section*{Acknowledgements}
This work was supported by the IITP grants (RS-2022-II220959: Few-Shot Learning of Causal Inference in Vision and Language for Decision Making (30\%), RS-2022-II220264: Comprehensive Video Understanding and Generation with Knowledge-based Deep Logic Neural Network (30\%), RS-2022-II220113: Developing a Sustainable Collaborative Multi-modal Lifelong Learning Framework (20\%), RS-2024-00457882: National AI Research Lab Project (20\%)) funded by the Ministry of Science and ICT, Korea.


%
%
\bibliographystyle{splncs04}
\bibliography{main}

\clearpage
\appendix
\section{Implementation Details}

\subsection{Overall}

\noindent{\textbf{Model Architecture.}}
We adopt Qwen3-VL~\cite{bai2025qwen3} as our language model backbone. For 3D scene understanding, we employ a pretrained Point3R~\cite{wu2025point3r} to reconstruct point clouds and maintain the memory representation throughout the generation. \\

\noindent{\textbf{Cognitive Map Representation.}}
Our Cognitive Map $\mathcal{M}_n = \{ (\mathbf{p}_{k}, \mathbf{f}_{k}, \mathbf{g}_{k}) \}_{k=1}^{K_n}$ integrates multiple feature modalities. 2D Visual features $\mathbf{f}_{k}$, extracted by the vision encoder of Qwen3-VL, are stored alongside the geometric features from Point3R memory. Additionally, since Qwen3-VL extracts deepstack features, we store these as supplementary visual representations. Specifically, the feature update function (Eq.~\ref{eq:feat_update}) is applied exclusively to visual features, excluding the deepstack features from this update process. \\

\noindent{\textbf{{Variable-Length Token Handling.}}
Since the Cognitive Map $\mathcal{M}_n$ contains a variable number of visual tokens $K_n$, we introduce a special visual token \texttt{<|pointer\_pad|>} that indicates 3D positions within the Cognitive Map. Following the video input format of Qwen3-VL, we sort the point tokens by timestep and insert text-based temporal markers between timesteps when processing multi-view images. \\

\noindent{\textbf{Positional Encoding.}}
We apply 1D-RoPE to the \texttt{<|pointer\_pad|>} tokens in the same manner as text tokens. Rather than employing 3D-RoPE for explicit spatial encoding, we leverage the geometric features provided by the Point3R encoder to implicitly capture 3D spatial information. \\

\noindent{\textbf{Training and Evaluation Setup.}}
During training, the vision encoder remains frozen to preserve pretrained visual representations. We train the language model backbone and the projector $\operatorname{Prj}(\cdot)$ with a learning rate of \texttt{1e-5}. For evaluation, we implement our model on the publicly available evaluation pipeline LLM-Eval~\cite{lmms_eval2024}. \\

\subsection{VSTI-Bench}

For evaluation on the VSTI benchmark~\cite{fan2025vlm3r}, we use the official training split provided by the official VSTI website.
During training, 32 images are used per sample.
A preprocessing stage first extracts semantic and geometric features from all 32 frames, and the resulting feature files are stored in advance.
The 32 frames are processed sequentially prior to the MLLM, as described in the main paper.
To improve training efficiency, the stored feature files are loaded directly during MLLM training.
Frame labels are placed before the visual tokens of each frame, following common practice in the Qwen2-VL setup.

\subsection{VSI-Bench}

In Table~\ref{tab:vstibench_full}, we compare our model against previous studies with 8 tasks proposed in VSI-Bench~\cite{vsibench}. 
During training, we conduct evaluation on VSI-Bench.
As VSI-Bench does not provide a training set, we utilize the training sets of both VLM-3R~\cite{fan2025vlm3r} and SPAR~\cite{zhang2025spar}.
Specifically, since VLM-3R does not include order-dependent questions, we additionally incorporate a 73K subset ($\sim$1\%) of SPAR-7M to cover such question types.
Unless otherwise specified, 32 frames are used for both training and evaluation.

\subsection{RoboFAC}

We explore the model behavior in three subsets of RoboFAC~\cite{robofac}: short-term, med-term, and long-term. 
Through this experiment, we demonstrate the token efficiency of our approach in robotic environments. 
We use Qwen3-VL-4B as the backbone network. Both our model and Qwen3-VL-4B are trained on the official training set of RoboFAC, sampling one frame per second following the practice of RoboFAC. Each frame is resized to $(512, 384)$, resulting in 192 visual tokens after patchifying and spatial merging. The number of frames given to the model varies from two frames to more than 50 frames. While the number of tokens for Qwen3-VL increases with the number of frames, our model generally keeps the number of tokens constant while preserving model performance.

\section{Comparison on 3D-LLM Benchmarks}

Following the evaluation of SR-3D, we also compare \methodname ~with 3D-LLM and video-LLM models on three popular 3D-LLM benchmarks, namely, ScanQA, SQA3D, and Scan2Cap. 
Motivated by two-stage straining employed in Video-3D-LLM, we fine-tune our model, pre-trained on the VSI-Bench training set, for respective dataset. 
We train the model for a single epoch while following the rest of hyperparameters with those used in training on VSI-Bench. 
In Table~\ref{tab:3d_general}, we report the results on ScanQA, SQA3D, and Scan2Cap using their respective metrics. We use the validation set for Scan2Cap and ScanQA, and the test set for SQA3D.

As shown in Table~\ref{tab:3d_general}, our \methodname achieves competitive performance with previous benchmarks although our approach does not rely on the GT point cloud.
Specifically, our model outperforms prior methods on SQA3D, while showing competitive or slightly lower performance on other benchmarks.
Due to the different properties of each benchmark, previous approaches also exhibit varying tendencies across datasets.
We note that directly leveraging pre-processed 3D point clouds limits practicability owing to the costly acquisition process.
Accordingly, approaches that directly perform 3D understanding from multi-view images, such as VG-LLM~\cite{zheng2025vgllm} and Spatial-MLLM~\cite{wu2025spatialmllm}, represent a promising direction for future research.

\begin{table}[!h]
    \caption{Evaluation of spatial scene understanding performance on the Scan2Cap, ScanQA, and SQA3D benchmarks. \textsuperscript{\dag} indicates methods evaluated in a zero-shot setting. } 
    \small
    \centering
    \resizebox{0.93\linewidth}{!}{
    
    \begin{tabular}{@{}lccccclclcccc@{}}
        \toprule
        & \multicolumn{5}{c}{ScanQA} & & \multicolumn{1}{c}{SQA3D} & & \multicolumn{4}{c}{Scan2Cap} \\
        \cmidrule(lr){2-6} \cmidrule(lr){8-9} \cmidrule(lr){10-13}
       Methods & B-4 $\uparrow$ & Rouge $\uparrow$ & Cider $\uparrow$ & Meteor $\uparrow$ & EM $\uparrow$ & & EM $\uparrow$ & & B-4 $\uparrow$ & Rouge $\uparrow$ & Cider $\uparrow$ & Meteor $\uparrow$ \\
        \midrule
    \rowcolor{navyblue!5}  
        \multicolumn{13}{l}{\bf\emph{Task-specific Specialist}}\\
        VoteNet+MCAN~\cite{yu2019deep}& 6.2 & 29.8 & 54.7 & 11.4 & 17.3 & & - &  & - & - & - & - \\
        ScanRefer+MCAN~\cite{yu2019deep} & 7.9 & 30.0 & 55.4 & 11.5 & 18.6 & & - & & - & - & - & - \\
        ScanQA~\cite{azuma2022scanqa} & 10.1 & 33.3 & 64.9 & 13.1 & 21.0 & & - &  & - & - & - & - \\
        3D-VisTA~\cite{zhu20233d} & 10.4 & 35.7 & 69.6 & 13.9 & 22.4 & & - &  & 34.0 & 54.3 & 66.9 & 27.1\\
        \midrule
    \rowcolor{navyblue!5}  
        \multicolumn{13}{l}{\bf\emph{2D Large Multi-modal Models}}\\
        Oryx-34B~\cite{liu2024oryx} & - & 37.3 & 72.3 & 15.0 & - & & - && - & - & - & - \\
        NaviLLM~\cite{zheng2024towards} & 12.0 & 38.4 & 75.9 & 15.4 & 23.0 & & - & & - & - & - & - \\
        LLaVA-Video-7B\textsuperscript{\dag}~\cite{zhang2024video} & 3.1 & 44.6 & 88.7 & 17.7 & - & & - & - & - & - & - &\\
        NaVILA~\cite{cheng2024navila} & 16.9 & 49.3 & 102.7 & 20.1 & 28.6 & & - &  & - & - & - & -  \\
        \midrule
    \rowcolor{navyblue!5}    
        \multicolumn{13}{l}{\bf\textit{3D Large Multi-modal Models}}\\
        Scene-LLM~\cite{fu2024scene} & 12.0 & 40.0 & 80.0 & 16.6 & 27.2 & & 54.2 && - & - & - & -  \\
        ChatScene~\cite{chatscene}  & 14.3 & 41.6 & 87.7 & 18.0 & 21.6 & & 54.6 && 36.3 & 58.1 & 77.2 & 28.0\\
        LLaVA-3D~\cite{llava3d} & 14.5 & \textbf{50.1} & 91.7 & \textbf{20.7} & 27.0 & & 55.6 & & 41.1 & 63.4 & 79.2 & \textbf{30.2} \\
        Video-3D LLM~\cite{video3dllm} & 16.2 & 49.0 & 102.1 & 19.8 & 30.1 & & 58.6 & & \textbf{42.4} & 62.3 & 83.8 & 28.9\\
        Spatial-MLLM-4B \cite{wu2025spatialmllm} & 14.8 & 45.0 & 91.8 & 18.4 & - && 55.9 & & - & - & - & - \\
        VG-LLM-8B~\cite{zheng2025vgllm}  &&-&-&-&-&-&&- & 41.5 & \textbf{62.6} & 80.0 & 28.9\\
        3DRS-7B~\cite{huang20253drs} & - & - & \textbf{104.8} & - & 30.3 && 60.6 && 41.6 & - & \textbf{86.1} & - \\
        \methodname-8B (Ours) & \textbf{17.0} & 49.8 & 102.8 & 19.9 & \bf31.3 & & \textbf{61.3} & & 40.8 & 62.3 & 79.4 & 28.4\\
        \hline
    \end{tabular}
    
    }
    \vspace{-0.2cm}
    \label{tab:3d_general}
\end{table}

\section{Additional Control Experiments}

\subsection{Effect of changing the number of frames}

In this work, we train and evaluate our \methodname using 32 frames provided by each dataset.
To demonstrate the effectiveness of our recurrent approach, we conduct additional experiments by evaluating the model trained with 32 frames across varying numbers of input frames.
This evaluation is feasible since our framework produces 3D memory states that avoid introducing redundant information even when different tokens are assigned to the same spatial location, thereby constructing a consistent 3D map regardless of the number of input frames.
Specifically, we evaluate on Scan2Cap~\cite{chen2021scan2cap} using 16, 32, 64, 128, 256, and 512 frames. 

\begin{table}[h]
\centering
\caption{Scan2Cap performance comparison across different numbers of input frames.}
\label{tab:scan2cap_frames}
\resizebox{0.98\linewidth}{!}{
\begin{tabular}{>{\raggedright\arraybackslash}p{0.2\linewidth}>{\centering\arraybackslash}p{0.2\linewidth}>{\centering\arraybackslash}p{0.2\linewidth}>{\centering\arraybackslash}p{0.2\linewidth}>{\centering\arraybackslash}p{0.2\linewidth}}
\toprule
\textbf{\# Frames} & \textbf{B-4} $\uparrow$ & \textbf{Rouge} $\uparrow$ & \textbf{Cider} $\uparrow$ & \textbf{Meteor} $\uparrow$ \\
\midrule
16  & 38.7 & 61.6 & 70.2 & 27.7 \\
32  & 40.8 & 62.3 & 79.4 & 28.4 \\
64  & 40.6 & 62.2 & 78.7 & 28.4 \\
128 & 40.0 & 61.9 & 76.3 & 28.1 \\
256 & 39.2 & 61.7 & 71.8 & 27.8 \\
512 & 38.5 & 61.5 & 69.2 & 27.5 \\
\bottomrule
\end{tabular}
}
\end{table}

\noindent As reported in Table~\ref{tab:scan2cap_frames}, our model achieves the best performance when evaluated with 32 frames, an identical setup to the training configuration.
Although the number of frames differs, our model maintains consistent performance, demonstrating robustness to variations in frame count.
This stems from the consistency of the 3D map regardless of how many images are provided as input.
Specifically, our model does not assign multiple tokens when observing the same region across different viewpoints.

\subsection{Using a static threshold $\delta$}

As described in Equation~\ref{eq:neighbor}, we employ a pre-defined threshold $\delta$ that is dynamically allocated based on the scene scale.
This design prevents an excessive number of tokens from being introduced due to large scene scales in outlier cases.
To validate the effectiveness of the dynamic threshold $\delta$, we compare our model against a variant that adopts a fixed $\delta = 0.2$ during the pre-processing stage.
We note that all hyperparameters are kept identical and only the token threshold is varied in this experiment.
Specifically, we compare them on Scan2Cap~\cite{chen2021scan2cap}, following other control experiments.

\begin{table}[h]
\centering
\caption{Comparison of Scan2Cap performance with static and dynamic thresholding.}
\label{tab:static_delta}
\resizebox{0.98\linewidth}{!}{
\begin{tabular}{>{\raggedright\arraybackslash}p{0.2\linewidth}>{\centering\arraybackslash}p{0.2\linewidth}>{\centering\arraybackslash}p{0.2\linewidth}>{\centering\arraybackslash}p{0.2\linewidth}>{\centering\arraybackslash}p{0.2\linewidth}}
\toprule
\textbf{Model} & \textbf{B-4} $\uparrow$ & \textbf{Rouge} $\uparrow$ & \textbf{Cider} $\uparrow$ & \textbf{Meteor} $\uparrow$ \\
\midrule
Static $\delta$  & 40.3 & 62.1 & 78.1 & 28.3 \\
Dynamic $\delta$ & 40.8 & 62.3 & 79.4 & 28.4 \\
\bottomrule
\end{tabular}
}
\end{table}

\noindent As reported in Table~\ref{tab:static_delta}, our model achieves slight improvement when using a dynamic threshold $\delta$.
Despite their minor performance gap, using dynamic $\theta$ enables much more stable training since for some scenes, the static $\theta$ introduces unnecessarily many tokens due to a large scene scale. 
Specifically, as the model observes different viewpoint, the scene scale could drastically increases. 
Therefore, we adopt the dynamic threshold throughout our framework.

\section{Full Results}

Owing to space constraints in the main paper, we provide the complete tables in this section.
In Table~\ref{tab:3d_vsibench_full}, we report the full results on VSI-Bench, encompassing large-scale LLM baselines.
Notably, our model retains state-of-the-art performance upon inclusion of large-scale LLM variants.
In Table~\ref{tab:vstibench_full}, we additionally report the full results on VSTI-Bench.

\begin{table}[!htbp]
    \caption{Results on multi-view global spatial scene understanding on VSI-Bench~\cite{yang2024thinking}. \textit{Spatial-Enhanced Models} denote methods specialized for spatial reasoning. \methodname achieves state-of-the-art overall performance. It performs particularly well on absolute distance, relative direction, and appearance order, demonstrating the benefits of explicit 3D representations for spatial reasoning. 
    }
    \label{tab:3d_vsibench_full}
    \vspace{-2.5mm}
    \centering
    \resizebox{1.00\linewidth}{!}{
    \begin{tabular}{l|c|cccccccc}
        & & 
        \rotatebox{30}{\textbf{Obj. Count}} &
        \rotatebox{30}{\textbf{Abs. Dist.}} &
        \rotatebox{30}{\textbf{Obj. Size}} & 
        \rotatebox{30}{\textbf{Room Size}} &
        \rotatebox{30}{\textbf{Rel. Dist.}} &
        \rotatebox{30}{\textbf{Rel. Dir.}} &
        \rotatebox{30}{\textbf{Route Plan}} &
        \rotatebox{30}{\textbf{Appr. Order}} \\
        \textbf{Model} & \textbf{Avg.} & \multicolumn{4}{c}{\cellcolor{orange!20}\textbf{Numerical Answer}} & \multicolumn{4}{c}{\cellcolor{yellow!20}\textbf{Multiple-Choice Answer}} \\
        \hline
        \rowcolor{navyblue!5}
        \hline
        \color[HTML]{969696}Random & \color[HTML]{969696}  & \color[HTML]{969696} & \color[HTML]{969696}- & \color[HTML]{969696}- & \color[HTML]{969696}- & \color[HTML]{969696}25.0 & \color[HTML]{969696}36.1 & \color[HTML]{969696}28.3 & \color[HTML]{969696}25.0 \\ 
        \color[HTML]{969696}Human Level\textsuperscript{\dag} & \color[HTML]{969696}79.2 & \color[HTML]{969696}94.3 & \color[HTML]{969696}47.0 & \color[HTML]{969696}60.4 & \color[HTML]{969696}45.9 & \color[HTML]{969696}94.7 & \color[HTML]{969696}95.8 & \color[HTML]{969696}95.8 & \color[HTML]{969696}100.0 \\
        \hline
        \rowcolor{navyblue!5}
    
        \multicolumn{1}{l|}{\textcolor{black}{\textit{Proprietary Models (API)}}} & & & & & & & & & \\
        GPT-4o~\cite{openai2024gpt4o} & 34.0 & 46.2 & 5.3 & 43.8 & 38.2 & 37.0 & 41.3 & 31.5 & 28.5 \\
        Gemini-1.5-Flash~\cite{team2024gemini} & 42.1 & 49.8 & 30.8 & 53.5 & {54.4} & 37.7 & 41.0 & 31.5 & 37.8 \\
        Gemini-1.5-Pro~\cite{team2024gemini} & 45.4 & {56.2} & {30.9} & 64.1 & 43.6 & {51.3} & 46.3 & {36.0} & 34.6 \\
        \hline
        \rowcolor{navyblue!5}
        \multicolumn{1}{l|}{\textcolor{black}{\textit{Open-source Models}}} & & & & & & & & & \\
        InternVL2-2B & 27.4 & 21.8 & 24.9 & 22.0 & 35.0 & 33.8 & {44.2} & 30.5 & 7.1 \\
        InternVL2-8B~\cite{chen2024internvl} & 34.6 & 23.1 & {28.7} & 48.2 & {39.8} & 36.7 & 30.7 & 29.9 & 39.6 \\
        InternVL2-40B~\cite{chen2024internvl} & 36.0 & 34.9 & 26.9 & 46.5 & 31.8 & 42.1 & 32.2 & 34.0 & 39.6 \\
        LongVILA-8B~\cite{xue2024longvila}  & 21.6 & 29.1 & 9.1 & 16.7 & 0.0 & 29.6 & 30.7 & 32.5 & 25.5 \\
        VILA-1.5-8B~\cite{lin2024vila} & 28.9 & 17.4 & 21.8 & 50.3 & 18.8 & 32.1 & 34.8 & 31.0 & 24.8 \\
        VILA-1.5-40B~\cite{lin2024vila}  & 31.2 & 22.4 & 24.8 & 48.7 & 22.7 & 40.5 & 25.7 & 31.5 & 32.9 \\
        LongVA-7B~\cite{zhang2024long} & 29.2 & 38.0 & 16.6 & 38.9 & 22.2 & 33.1 & 43.3 & 25.4 & 15.7 \\
        LLaVA-NeXT-Video-7B~\cite{zhang2024llavanextvideo} & 35.6 & 48.5 & 14.0 & 47.8 & 24.2 & {43.5} & 42.4 & 34.0 & 30.6 \\
        LLaVA-NeXT-Video-72B~\cite{zhang2024llavanextvideo}  & 40.9 & {48.9} & 22.8 & 57.4 & 35.3 & 42.4 & 36.7 & {35.0} & 48.6 \\
        LLaVA-OneVision-0.5B~\cite{li2025llavaonevision}  & 28.0 & 46.1 & 28.4 & 15.4 & 28.3 & 28.9 & 36.9 & 34.5 & 5.8 \\
        LLaVA-OneVision-7B~\cite{li2025llavaonevision}  & 32.4 & 47.7 & 20.2 & 47.4 & 12.3 & 42.5 & 35.2 & 29.4 & 24.4 \\
        LLaVA-OneVision-72B~\cite{li2025llavaonevision} & 40.2 & 43.5 & 23.9 & {57.6} & 37.5 & 42.5 & 39.9 & 32.5 & 44.6 \\
        \hline
        \rowcolor{navyblue!5}
        \multicolumn{1}{l|}{\textcolor{black}{\textit{Spatial-Enhanced Models}}} & & & & & & & & & \\
        SAT-LLaVA-Video-7B~\cite{ray2025satdynamicspatialaptitude} & - & - & - & - & 47.3 & 41.1 & 37.1 & 36.1 & 40.4 \\
        SPAR-8B~\cite{zhang2025flatland} & 41.1 & - & - & - & - & - & - & -  & - \\
        VG-LLM-4B~\cite{zheng2025vgllm} & 47.3 & 66.0 & 37.8 & 55.2 & 59.2 & 44.6 & 45.6 & 33.5 & 36.4 \\
        Spatial-MLLM-4B~\cite{wu2025spatialmllm} & 48.4 & 65.3 & 34.8 & 63.1 & 45.1 & 41.3 & 46.2 & 33.5 & 46.3 \\
        VG-LLM-8B~\cite{zheng2025vgllm} & 50.7 & 67.9 & 37.7 & 58.6 & 62.0 & 46.6 & 40.7 & 32.4 & 59.2\\
        3DRS-7B~\cite{huang20253drs} & 45.9 & 68.7 & 34.8 & 53.6 & 56.6 & 40.9 & 43.2 & 30.4 & 39.2 \\
        VLM-3R-7B~\cite{fan2025vlm3r} & \third{60.9} & \second{70.2} & \second{49.4} & \second{69.2} & \second{67.1} & \first{65.4} & \second{80.5} & \first{45.4} & \third{40.1} \\
        VST-7B~\cite{yang2025vst} & \second{61.2} & \first{71.6} & \third{43.8} & \first{75.5} & \first{69.2} & \third{60.0} & \third{55.6} & \second{44.3} & \first{69.2} \\
        \methodname-8B (ours) & \first{65.1} & \third{69.6} & \first{54.8} & \third{67.8} & \second{67.1} & \second{64.8} & \first{85.6} & \third{43.0} & \second{67.9} \\
        \bottomrule
    \end{tabular}
    }
    \vspace{-3.5mm}
\label{tab:vsibench_full}
\end{table}
\begin{table}[!htbp]
    \caption{
        Performance comparison on VSTI-Bench~\cite{fan2025vlm3r}, which evaluates joint spatial and temporal understanding. \textsuperscript{\dag} indicates methods tested on the \texttt{Tiny} subset. 
        \methodname achieves strong performance on spatial reasoning and camera movement prediction tasks, demonstrating its ability to encode both geometric and temporal cues within a unified 3D representation.
    }
    \label{tab:vstibench_full}
    \vspace{-2.5mm}
    \centering
    \resizebox{1.00\linewidth}{!}{
    \begin{tabular}{l|c|ccccc}
        & &
        \rotatebox{30}{\textbf{Cam-Obj. Dist.}} &
        \rotatebox{30}{\textbf{Cam. Displce.}} &
        \rotatebox{30}{\textbf{Cam. Mov.}} &
        \rotatebox{30}{\textbf{Obj-Obj. Pose}} &
        \rotatebox{30}{\textbf{Cam-Obj. Dist.}} \\
        \textbf{Model} & \textbf{Avg.} &
        \multicolumn{2}{c}{\cellcolor{orange!20}\textbf{Numerical Answer}} &
        \multicolumn{3}{c}{\cellcolor{yellow!20}\textbf{Multiple-Choice Answer}} \\
        \hline
        \rowcolor{navyblue!5}
        \hline
        \color[HTML]{969696}Random & \color[HTML]{969696}- & \color[HTML]{969696}- & \color[HTML]{969696}- & \color[HTML]{969696}36.1 & \color[HTML]{969696}50.0 & \color[HTML]{969696}36.1 \\
        \color[HTML]{969696}Frequency & \color[HTML]{969696}27.4 & \color[HTML]{969696}5.4 & \color[HTML]{969696}6.2 & \color[HTML]{969696}40.7 & \color[HTML]{969696}52.2 & \color[HTML]{969696}32.4 \\
        \color[HTML]{969696}Human Level\textsuperscript{\dag} & \color[HTML]{969696}77.0 & \color[HTML]{969696}51.4 & \color[HTML]{969696}46.8 & \color[HTML]{969696}95.1 & \color[HTML]{969696}97.5 & \color[HTML]{969696}94.3 \\
        \hline
        \rowcolor{navyblue!5}
        \multicolumn{1}{l|}{\textcolor{black}{\textit{Proprietary Models (API)}}} & & & & & & \\
        Gemini-1.5-Flash~\cite{team2024gemini} & 32.1 & 28.5 & 20.9 & 24.4 & 52.6 & 33.9 \\
        GPT-4o~\cite{openai2024gpt4o} & 38.2 & 29.5 & 23.4 & 37.3 & 58.1 & 42.5 \\
        \hline
        \rowcolor{navyblue!5}
        \multicolumn{1}{l|}{\textcolor{black}{\textit{Open-source Models}}} & & & & & & \\
        LLaVA-OneVision-0.5B~\cite{li2025llavaonevision} & 36.9 & 16.5 & \third{32.4} & 46.1 & 50.5 & 39.0 \\
        InternVL2-2B~\cite{chen2024internvl} & 38.1 & 17.7 & 27.8 & 43.0 & 54.9 & 47.2 \\
        LongVILA-8B~\cite{xue2024longvila} & 30.5 & 20.0 & 11.6 & 35.4 & 52.3 & 33.4 \\
        LongVA-7B & 32.3 & 13.5 & 5.1 & 43.7 & 57.9 & 41.2 \\
        VILA-1.5-8B~\cite{lin2024vila} & 37.3 & 30.1 & 27.3 & 42.2 & 50.4 & 36.7 \\
        LLaVA-NeXT-Video-7B~\cite{liu2024llavanext} & 40.0 & 28.2 & 1.8 & \third{49.8} & 64.7 & \third{55.6} \\
        LLaVA-OneVision-7B~\cite{li2025llavaonevision} & 41.7 & 29.9 & 19.3 & 47.5 & 62.1 & 49.8 \\
        InternVL2-8B~\cite{chen2024internvl} & 43.5 & \third{32.9} & 13.5 & 48.0 & 68.0 & 55.0 \\
        VILA-1.5-40B~\cite{lin2024vila} & 38.2 & 28.2 & 15.7 & 28.8 & 65.4 & 53.0 \\
        LLaVA-NeXT-Video-72B~\cite{liu2024llavanext} & \third{44.0} & 32.3 & 10.5 & 48.1 & \third{78.3} & 50.9 \\
        \hline
        \rowcolor{navyblue!5}
        \multicolumn{1}{l|}{\textcolor{black}{\textit{Spatial-Enhanced Models}}} & & & & & & \\
        VLM-3R-7B~\cite{fan2025vlm3r} & \second{58.8} & \second{39.4} & \second{39.6} & \second{60.6} & \second{86.5} & \second{68.6} \\
        \methodname-8B (ours) & \first{67.5} & \first{40.9} & \first{47.1} & \first{88.1} & \first{90.9} & \first{70.6} \\
        \bottomrule
    \end{tabular}
    }
\vspace{-3.5mm}
\end{table}

\clearpage
\section{Qualitative Results}

In this section, we provide additional qualitative results demonstrating the effectiveness of our proposed \methodname.
In Figure~\ref{fig:attention_maps_scan2cap_1} and~\ref{fig:attention_maps_scan2cap_2}, we visualize the reconstructed pointmap from multi-view images in Scan2Cap~\cite{chen2021scan2cap} alongside attention maps obtained by varying the text query.
Although our model is not explicitly supervised to attend to the target object, the 3D cognitive map facilitates visual understanding of MLLMs through an explicit 3D representation.
As evidence, the attention map highly concentrates on the target object.
We also visualize the qualitative results on RoboFAC~\cite{robofac} in Figure~\ref{fig:robofac_attn}.
The attention maps demonstrate that our model accurately concentrates on the target objects relevant to pick-and-place operations in the robotics environment, validating the interpretability of our approach.

\begin{figure}[!htbp]
    \centering
    \includegraphics[width=1.0\linewidth]{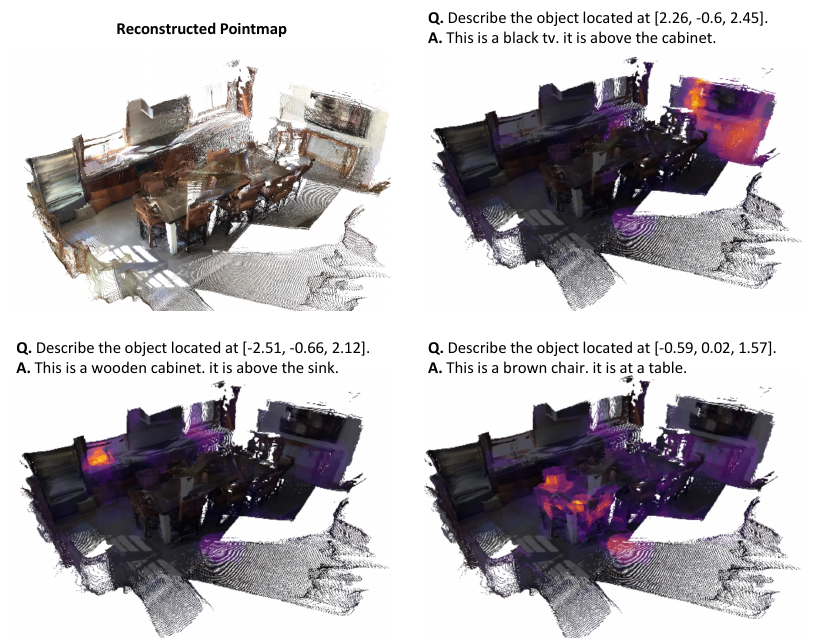}
    \caption{
        Visualization of attention scores over visual tokens on a validation sample from Scan2Cap~\cite{chen2021scan2cap}. \methodname assigns high attention scores to the visual tokens corresponding to the generated answer.
    }
    \label{fig:attention_maps_scan2cap_1}
\end{figure}

\begin{figure}[!htbp]
    \centering
    \includegraphics[width=1.0\linewidth]{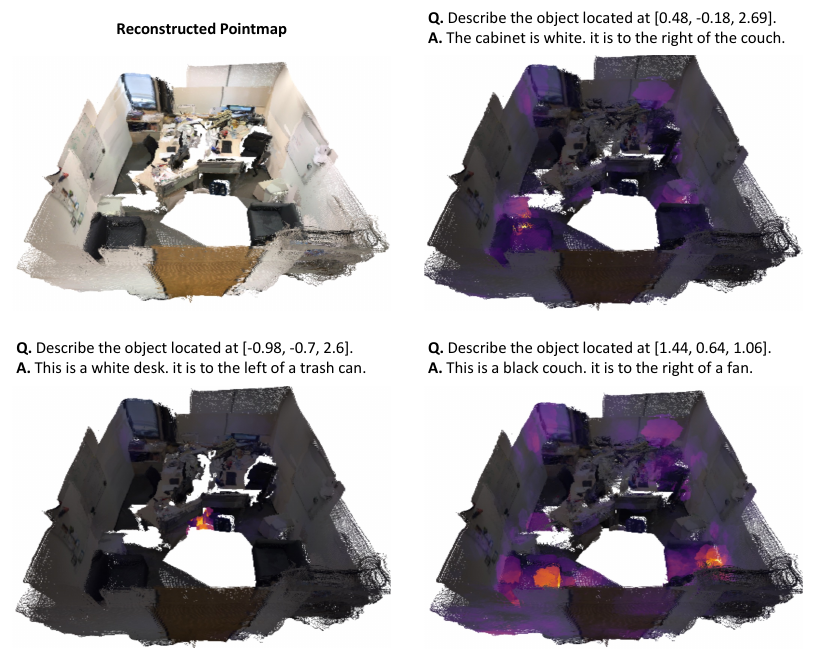}
    \caption{
        Visualization of attention scores over visual tokens on a validation sample from Scan2Cap~\cite{chen2021scan2cap}. \methodname assigns high attention scores to the visual tokens corresponding to the generated answer.
    }
    \label{fig:attention_maps_scan2cap_2}
\end{figure}

\begin{figure}[!htbp]
    \centering
    \includegraphics[width=1.0\linewidth]{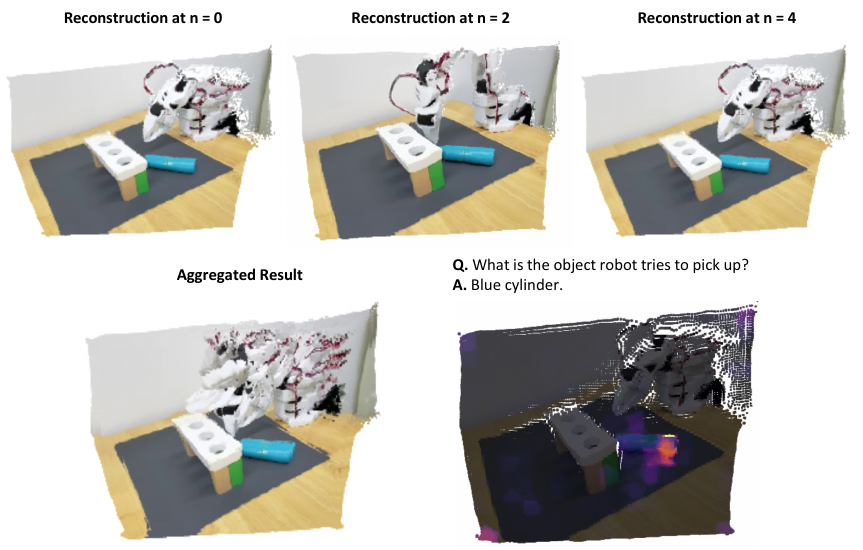}
    \caption{
    Reconstructed results of a sample video from RoboFAC~\cite{robofac} and its attention map visualization. \methodname aggregates visual tokens across timesteps while preserving their temporal order, maintaining the spatial layout of the scene and the trajectory of moving objects.
    }
    \label{fig:robofac_attn}
\end{figure}

\clearpage

\end{document}